%% file: main.tex
\pdfoutput=1 % ensure pdflatex (for arXiv)

\documentclass[11pt]{article}
\usepackage[final]{acl}
\usepackage{nert}
\usepackage{times}
\usepackage{latexsym}
\usepackage[inline,shortlabels]{enumitem}

\usepackage{microtype}

\usepackage{kotex}

\newcounter{mycounter}

\newcommand\labelledmodelcounter[1]{\refstepcounter{mycounter}\textbf{(\themycounter)}\label{model:#1}}
\newcommand{\modelref}[1]{\textbf{(\ref{model:#1})}}

\title{Adapting BigScience Multilingual Model to Unseen Languages}
\input{sections/0_authors}

\begin{document}
\maketitle
\input{sections/09_abstract}

\input{sections/10_introduction}

\begin{figure*}[ht]
\includegraphics[width=12cm]{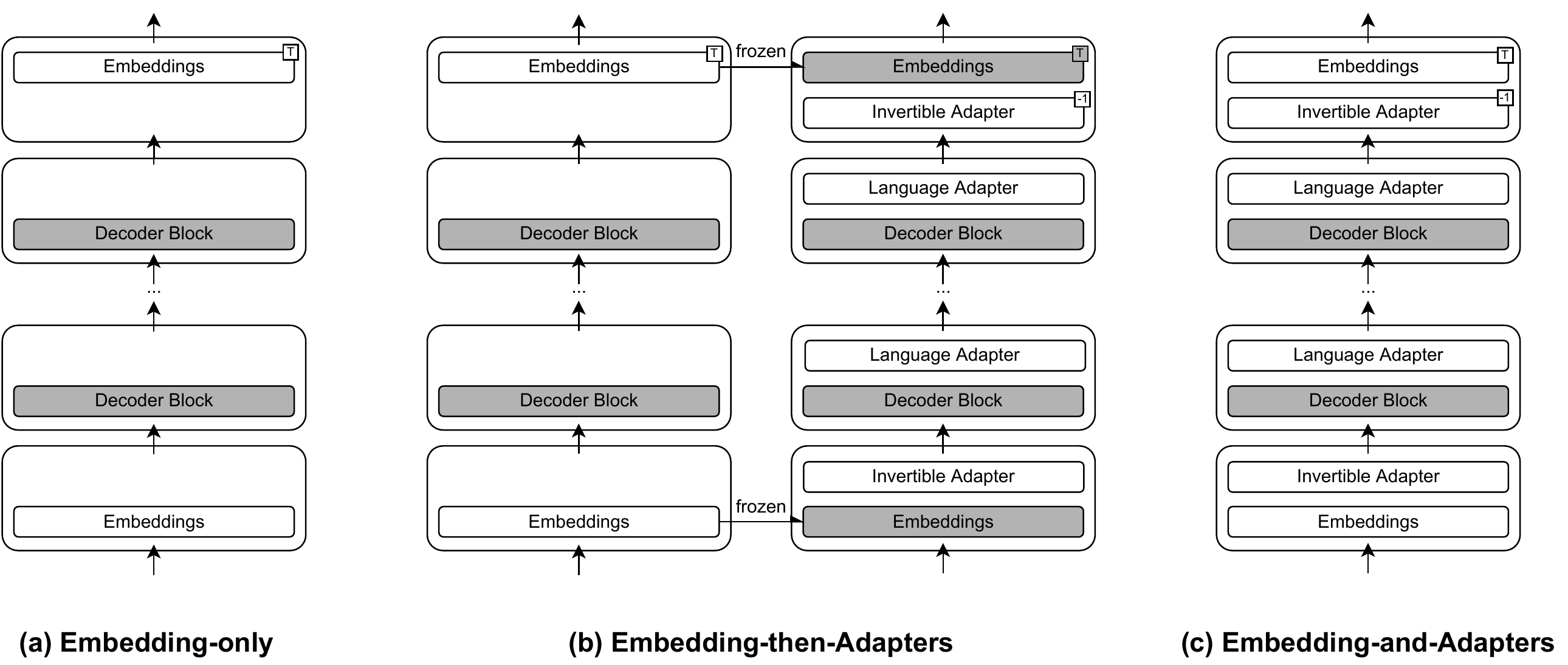}
\centering
\caption{Three different language adaptation strategies: (a) Embedding-only strategy: finetune the embedding layer of the pretrained model, which is tied with the language modeling head. (b) Embedding-then-Adapters strategy: first finetune the embedding layer, then train the invertible adapter and language adapters. (c) Embedding-and-Adapters strategy: finetune the embedding layer, invertible and language adapters at the same time. Grey blocks indicate frozen layers whereas white blocks indicate trainable layers.}
\label{fig:language-adaptation}

\end{figure*}

\input{sections/20_related_work}

\begin{table*}[ht]
\small
    \centering
    \begin{tabular}{lllll}
        \toprule
        \multirow{2}{*}{Language} & \multirow{2}{*}{Prompt Template} & \multicolumn{3}{c}{Candidate Verbalizer for \texttt{[MASK]}}\\
        % \cmidrule{3-5} \\
        {} & {} & Entailment & Contradiction & Neutral \\
        \midrule
        en & \texttt{[premise]}, right? \texttt{[Mask]}, \texttt{[hypothesis]} & Yes & No & Also \\
        de & \texttt{[premise]}, richtig? \texttt{[Mask]}, \texttt{[hypothesis]} & Ja & Nein & Auch \\
        ko & \texttt{[premise]}, 맞지? \texttt{[Mask]}, \texttt{[hypothesis]} & 예 & 아니요 & 또한 \\
        \bottomrule
    \end{tabular}
    \caption{Natural language inference (NLI) prompts in English, German (de) and Korean (ko). The German and Korean prompts are manually translated from the English counterpart.}
    \label{tab:zero-shot-prompt-template}
\end{table*}

\begin{table*}[ht]
\small
    \centering
    \begin{tabular}{lllllcc}
        \toprule
        Dataset & Language & $|\text{Train}|$ & $|\text{Dev}|$ & $|\text{Test}|$ & Avg Len Prem & Avg Len Hyp \\
        \midrule
        XNLI \cite{conneau-etal-2018-xnli} & en & 392,702 & 2,490 & 5,010 & 21.7 & 10.7 \\
        XNLI \cite{conneau-etal-2018-xnli} & de & 392,702 & 2,490 & 5,010 & 21.1 & 10.8 \\
        KLUE-NLI \cite{park-2021-klue} & ko & 24,998 & 3,000 & 3,000 & 47.2 & 25.5 \\
        \bottomrule
    \end{tabular}
    \caption{Statistics for NLI datasets in English (en), German (de), and Korean (ko). The three columns after the "Language" column reports the number of sentence pairs in train, dev, and test sets. "Avg Len Prem" and "Avg Len Hyp" columns refer to the average number of tokens in the premise and hypothesis sentences, respectively}
    \label{tab:nli-task}
\end{table*}

\begin{table*}[ht]
    \centering
    \small

    \begin{tabular}{llp{1.5cm}p{1cm}p{1.4cm}p{1cm}p{0.6cm}p{0.6cm}p{0.6cm}p{0.6cm}p{0.6cm}p{0.6cm}}
        \toprule
        & \textbf{Models} & \textbf{Strategies} & \textbf{Ckpt.} & \textbf{Emb.} & \textbf{Adpt. Red.} & \textbf{(p.) de} & \textbf{en$\rightarrow$ de} & \textbf{de$\rightarrow$ de} & \textbf{(p.) ko} & \textbf{en$\rightarrow$ ko} & \textbf{ko$\rightarrow$ ko} \\
        \midrule
        \labelledmodelcounter{mbert} & $\text{mBERT}_{BASE}$ & - & - & - & - & - & 70.0 & 75.5 & - & 69.7 & 72.9\\
        \labelledmodelcounter{xlmr} & $\text{XLMR}_{LARGE}$ & - & - & - & - & - & 82.5 & 85.4 & - & 80.4 & 86.4\\
        \labelledmodelcounter{xglm} & $\text{XGLM}_{1.7B}$ & - & - & - & - & 45.4 & - & - & 45.17 & - & - \\
        \midrule
        \labelledmodelcounter{bs} & BigScience & - & - & - & - & 34.1 & 44.8 & 67.4 & - & - & - \\
        \labelledmodelcounter{bs_emb} & BigScience & Emb & 118,500 & wte,wpe & - & 41.4 & 50.7 & 74.3 & 34.4 & 45.6 & 53.4 \\
        \labelledmodelcounter{bs_emb1_adapt2} & BigScience & Emb$\rightarrow$Adpt & 118,500 & wte,wpe & 16 & 40.0 &	50.5 & 69.9 & 33.8 & 40.4 & 51.8\\
        \labelledmodelcounter{bs_emb_adapt} & \textbf{BigScience} & \textbf{Emb+Adpt} & \textbf{118,500} & \textbf{wte} & \textbf{16} & \textbf{42.4} & \textbf{58.4} & \textbf{73.3} & \textbf{38.8} & \textbf{49.7} & \textbf{55.7} \\
        \midrule
        \labelledmodelcounter{bs_emb_adapt48} & BigScience & Emb+Adpt & 118,500 & wte & 48 & 42.4 & 57.6 & 73.7 & 36.3 & 48.3 & 52.9\\
        \labelledmodelcounter{bs_emb_adapt384} & BigScience & Emb+Adpt & 118,500 & wte & 384 & 42.4 & 55.3 & 74.2 & 37.5 & 49.4	& 54.6\\
        \midrule
        \labelledmodelcounter{bs100500_emb_adapt} & BigScience & Emb+Adpt & 100,500 & wte & 16 & 44.3 & 56.9 & 73.2 & 37.5 & 48.6 & 50.8\\
        \labelledmodelcounter{bs12000_emb_adapt} & BigScience & Emb+Adpt & 12,000 & wte & 16 & 33.5 & 55.2 & 70.5 & 32.9 & 46.4 & 53.3 \\
        \midrule
        \labelledmodelcounter{bs_ko_embpe_adapt} & BigScience & Emb+Adpt & 100,500 & wte,wpe & 16 & - & - & - & 37.5 & 53.5 & 63.5 \\
        \labelledmodelcounter{bs_de_embpe_adapt} & BigScience & Emb+Adpt & 118,500 & wte,wpe & 16 & 44.7 &	64.9 & 73.0 & - & - & -\\
        \bottomrule
    \end{tabular}
    \caption{Evaluation of language adaptation for German (de) and Korean (ko) on NLI with three baselines ($\text{mBERT}_{BASE}$ \cite{devlin-etal-2019-bert}, $\text{XLMR}_{BASE}$ \cite{conneau-etal-2020-unsupervised}, $\text{XGLM}_{1.3B}$ \cite{xi-2021-few-shot}). "Strategies" column indicates language adaptation strategies, which cover Embedding-only (Emb), Embedding-then-Adapters (Emb$\rightarrow$Adpt) and Embedding-and-Adapters (Emb+Adpt). "Ckpt." column stands for the BigScience pretrained checkpoint, "Emb." column the types of embedding layers (wte: token embedding, wpe: positional embedding), and "Adpt. Red." column the reduction factor for language adapters. "(p.) {de/ko}" column reports the prompt-based zero-shot evaluation result, "en$\rightarrow${de/ko}" cross-lingual result, and "{de/ko}$\rightarrow${de/ko}" supervised finetuning result. Row~\modelref{bs_emb_adapt} is bolded as all other language adaptation strategies and design choices are compared against it.}
    \label{tab:nli_result}
\end{table*}

\input{sections/30_methods}

\input{sections/40_results}

\input{sections/50_limitations}

\input{sections/acknowledgments}

\bibliography{main,anthology}

\appendix
\input{sections/100_appendix}

\end{document}

%% file: sections/0_authors.tex
\author{Zheng-Xin Yong\thanks{\,\, Both authors contributed equally.} \\
Brown University \\
\texttt{zheng\_xin\_yong@brown.edu}
\\\And 
Vassilina Nikoulina$^*$ \\ NAVER LABS Europe \\ \texttt{vassilina.nikoulina@naverlabs.com }}

%% file: sections/09_abstract.tex
\begin{abstract}
% We benchmark different ways of adpating  German and Korean languages into the autoregressive multilingual language model pretrained by the BigScience project, which supports 13 languages.  We investigate the best language adaptation strategy and the potential costs one would have to face if needed to add a new language to this model, and the trade-offs between the computational costs and expected performance. 

We benchmark different strategies of adding new languages (German and Korean) into the BigScience's pretrained  multilingual language model with 1.3 billion parameters that currently supports 13 languages. We investigate the factors that affect the language adaptability of the model and the trade-offs between computational costs and expected performance.
\end{abstract}

%% file: sections/10_introduction.tex
\section{Introduction}
Pretrained multilingual language models (LMs) have enabled cross-lingual transfer \cite{artetxe-etal-2020-cross, conneau-etal-2020-emerging, k-etal-2021-analyzing}. While several works suggest that such knowledge transfer goes beyond just vocabulary sharing across languages  \cite{artetxe-etal-2020-cross, k-etal-2021-analyzing}, others have shown that the models' performance is sensitive to the quality of the tokenization it relies on \cite{pfeiffer-etal-2021-unks}. Moreover, the quality of knowledge transfer can  degrade for unseen languages, especially when the scripts are unknown to the model \cite{muller-etal-2021-unseen}. 
 %for languages that are well-represented in the pretraining data or unseen languages that share vocabulary items with seen languages \cite{artetxe-etal-2020-cross,ebrahimi-kann-2021-adapt,lauscher-etal-2020-zero}. 
 
Data availability, but also the \textit{curse of multilinguality} \cite{conneau-etal-2020-unsupervised} makes training a single model covering all the languages challenging: there is a trade off between the number of languages in pretraining data, model capacity and the downstream performance for each individual language. Finally, it is  hard to anticipate any potential usage of pretrained LM in advance, hence it is important to study \textit{a posteoriori} adaptation to new tasks/languages.  
Recently proposed methods include \textbf{continual pretraining} of the model (restricted to the embedding layer training only in some cases) \cite{artetxe-etal-2020-cross,chau-etal-2020-parsing,muller-etal-2021-unseen,zhang-etal-2020-multi-stage,wang-etal-2020-extending}, or training of \textbf{language-specific adapters} \cite{pfeiffer-etal-2020-mad, pfeiffer-etal-2021-adapterfusion, pfeiffer-etal-2021-unks, philip-etal-2020-monolingual, ustun-etal-2021-multilingual, berard-2021-continual} for the target language. The core motivation behind these methods is to benefit from knowledge transfer encoded in the pretrained LM for the new language processing at a small computational cost (compared to full model retraining) . 

In this work, we aim at better understanding the trade-offs between the amount of compute and the final downstream task performance. Specifically, we study the impact of the following three factors: original pretraining steps, adaptation strategies, and adapters' capacity on the Natural Language Inference (NLI) task. As part of the initiative of BigScience, we experiment with its multilingual LM, which currently only supports 13 languages.  We investigate how researchers could adapt BigScience's full-open-access multilingual models to their languages of interest. Moreover, BigScience open source its intermediate pretraining checkpoints, therefore we also study how does the adaptability to new language changes with the amount of pretraining. 
%so we can study the effect of pretraining steps on language adaptation.

%\cite{Lu2021_unversal} has shown that representations learnt by a transformer model pretrained on English texts can be transferred on other modalities. 

Our main findings are: (1) the most promising strategy is training of the embedding layers and adapters simultaneously; (2) position encoding is an important component to adapt; (3) pretraining steps benefit zero-shot performance; (4) adaptation to low-resource languages can be done with low parameter budget.

%% file: sections/20_related_work.tex
\section{Background and Related Work}
Most of the works investigating extension of pretrained models to new languages consider models such as BERT \cite{devlin-etal-2019-bert} and XLM-R \cite{conneau-etal-2020-unsupervised} that were pretrained to 100+ languages with Masked Language Modeling (MLM) objective. 
\citet{artetxe-etal-2020-cross, pfeiffer-etal-2021-unks} demonstrate that it is possible to add new language to these models by training new embedding layer. \citet{muller-etal-2021-unseen} continues training the pretrained mBERT on the new language data, and finds that transliteration of languages using non-latin script boosts performance on these languages.

\citet{berard-2021-continual} adds new languages into pretrained multilingual NMT model by training embedding layer and adapter layers. It shows that higher adaptation cost is required for new target languages (as opposed to new source languages). 

%\cite{arr_jan2022} proposes to revise the architecture of pretrained multilingual models by adding an explicit language-specific module. They demonstrate that such architecture suffers less from \textit{curse of multilinguality} and allows to easily add new language to the pretrained model by training new language-specific module. 

Closest to our work \cite{ebrahimi-kann-2021-adapt} studies different approaches (i.e., continued pretraining, vocabulary expansion and adapter layers) allowing to extend XLM-R model to 1600 languages. They conclude that continued pretraining is the most promising direction. However, the cost of such pretraining will grow with the size of the pretrained model and can be quite prohibitive for many researchers working with low-resource languages. 

Our work aims to compare lightweight strategies  of adaptation to new language (embedding layer training and adapter layers) across the following previously unstudied dimensions: we consider the (1) adaptation parameter budget and (2) intermediate pretraining checkpoints by analyzing how they impact autoregressive model's adaptability to new languages.

% \begin{itemize}
%     % \item we study the language adaptation for multilingual autoregressive models;
%     \item we consider the adaptation parameter budget and analyse how it impacts model's performance;
%     \item we consider intermediate pretraining checkpoints and study how the pretraining amount of steps impacts model's adaptability to new languages. 
% \end{itemize}

%% file: sections/30_methods.tex
\section{Experimental settings}
\subsection{BigScience Pretrained Multilingual Language Model} 
BigScience open-sources one of its pretrained multilingual autoregressive LMs\footnote{https://huggingface.co/bigscience/tr5b-1B3-multilingual-alpha-checkpoints/tree/main} \cite{bigscience_engineering} that relies on transformer architecture with 24 decoder layers, 16 attention heads, embedding dimension of 2048, feed forward layer with dimensionality of 8192, and 1.3B parameters. This model is trained with Next Word Prediction objective (\textit{aka} GPT-style). It is pretrained on OSCAR deduplicated subcorpora \citep{ortiz:oscar_2019} for the following thirteen languages for a total of training 400B tokens: Indonesian, Basque, Vietnamese, Chinese, Urdu, Spanish, Catalan, Portuguese, French, English, Hindi, Arabic, and Bengali. See Appendix~\ref{appendix:bigs-pretraining} for further model pretraining details. 

In our experiments, we use the model checkpoint saved after 12,000, 100,500, and 118,500 pretraining steps (final checkpoint).

\subsection{New Languages and Data}
The need for adaptation of pretrained multilingual LM to new languages typically raises for low-resource languages for three reasons: first, the lack of the available data for such languages; second, the lack of the computational resources which people working on low-resource languages may experience; lastly, the lack of the evaluation resources, which makes it hard to assess the quality of the adapted model properly. In this study we simulate low-resource settings by restricting the amount of training samples for high-resource languages. This allows us to benefit from good-quality evaluation datasets available for the high resource languages. We expect that  our observations could generalize to real low-resource scenarios. 
We've chosen two languages for the adaptation of BigScience model:
    (1) \textit{German} (easy language) that belongs in the same Germanic language family as English and shares the Latin script with several pretraining languages; (2) \textit{Korean} (difficult language) that belongs to Koreanic language family and uses Hangul script system, that none of the languages of pretrained LM cover.
%We choose to adapt the model to \textbf{German} and \textbf{Korean} because the former is linguistically similar to languages seen in pretraining whereas the latter is not. German . On the other hand, none of the languages in pretraining data belong to the Koreanic language family and Hangul script system.

We train the pretrained model's byte-level BPE tokenizer with the vocabulary size of 130,000 on German and Korean deduplicated OSCAR subcorpora \citep{ortiz:oscar_2019}, which has 21.5B and 1.1B words respectively. Then, we use 100,000 samples from OSCAR subcorpora of the respective languages for adaptation.

\subsection{Language Adaptation Strategies}
We experiment with the following three language adaptation strategies that involve finetuning of the embedding layer and adapters layers (see Figure~\ref{fig:language-adaptation}). See Appendix~\ref{appendix:language-adaptation-training} for further training details.

\paragraph{Embedding-only.} The embedding layer consists of token embeddings and positional embeddings. We follow \citet{artetxe-etal-2020-cross} by learning new token embeddings and positional embeddings while freezing the rest of the transformer parameters. % As our preliminary result (Table~\ref{tab:nli_result}) indicates the benefits of continually pretraining both embeddings.
%we follow \citet{artetxe-etal-2020-cross} by learning new token embeddings and positional embeddings while freezing the rest of the transformer parameters. We use the same autoregressive pretraining objective for continual training with 100,000 steps and the batch size of 8 on a single V100 GPU. 
\vspace{-0.2cm}
\paragraph{Embedding-then-Adapters.} Following the Embedding-only approach, we first train the embedding layer on the target language pretraining data for 25,000 steps. Then, we freeze the entire transformer and add MAD-X invertible and language adapters \cite{pfeiffer-etal-2020-mad}. We subsequently finetune the adapters for an additional 25,000 steps.
% \vspace{-0.2cm}
\paragraph{Embedding-and-Adapters.} Instead of training the embedding layer and the adapters separately, we train both of them at the same time for 50,000 steps. We also vary the adapter capacity, which is defined by the reduction factor (also known as compression rate \cite{ruckle-etal-2021-adapterdrop}) in the adapter’s bottleneck layer. A smaller reduction value would lead to a larger amount of adapter parameters and, accordingly, a larger adaptation capacity. We consider the reduction factors of 16 (default), 48, and 384.

\subsection{Evaluation}
We evaluate our models on the natural language inference (NLI) task, using XNLI dataset \cite{conneau-etal-2018-xnli} for German and KLUE-NLI dataset \cite{park-2021-klue} for Korean\footnote{We evaluate our models on the dev set of KLUE-NLI.}. Table~\ref{tab:nli-task} reports the statistics of the NLI datasets.

We adopt the three following evaluation settings with increasing levels of task supervision. 
% \vspace{-0.2cm}
\paragraph{Prompt-based Zero-shot} Following \citet{xi-2021-few-shot}, we use the German and Korean cloze-style prompt templates (see Table~\ref{tab:zero-shot-prompt-template}). The zero-shot prediction is the candidate label verbalizer, which replaces the \texttt{[MASK]} token, that maximizes the LM based likelihood of the prompt sentence. 
% \vspace{-0.2cm}
\paragraph{Cross-lingual} 
%For the embedding-only approach, we follow \citet{pmlr-v119-hu20b} by finetuning the final classification head of the model on the labelled English training data in XNLI and subsequently evaluating it on the target language. 
In this evaluation setting, we follow \citet{pfeiffer-etal-2020-mad} and train task adapters on English task dataset. These adapters are then used on the target language (German or Korean) evaluation set. We adopt adapter size (reduction factor = 16) and run training for 2 epochs with the batch size of 32,  learning rate of $5e-5$, and maximum sequence length of 128 input tokens on a single V100 GPU, which takes around 5 hours to complete and produces around 0.65 kg of $\text{CO}_2$.

\paragraph{Supervised Finetuning} Our finetuning strategies are identical to the cross-lingual evaluation setting above, except that the classification head and the task adapters are finetuned on the target language dataset. Note that this implies that we train different task adapter for each model.

%% file: sections/40_results.tex
\section{Results and Discussion}
Table \ref{tab:nli_result} reports the results on different adaptation strategies. 
\paragraph{Baselines.} XLM-R and mBERT are pretrained multilingual LMs that were trained on 100+ languages and include both German and Korean languages. Therefore the adaptation of these models can be seen as an upper bound performance one could achieve on classification tasks. 
XGLM \cite{xi-2021-few-shot} is a multilingual autoregressive LM that was pretrained on 30 languages, including both German and Korean. It serves as the (upper-bound) baseline for the prompt-based evaluation setting due to its zero-shot learning capability through prompts on NLI \cite{xi-2021-few-shot}.

\paragraph{Language Adaptation Strategies.}

First, original BigScience multilingual model~\modelref{bs} performs poorly on German\footnote{We did not perform evaluation on Korean as the vocabulary used by BigScience model doesn't cover Hangul at all.}. Training adapter layers \modelref{bs_emb_adapt}, \modelref{bs_emb1_adapt2} boosts performance compared to the model updating embedding layers only \modelref{bs_emb}, especially under the cross-lingual setting for both German and Korean languages. Model \modelref{bs_emb1_adapt2} that trains embedding layers and adapters separately seems to perform worse than models training them simultaneously (models~\modelref{bs_emb_adapt} and~\modelref{bs_de_embpe_adapt}). 
 
 %, ~Table~\ref{tab:nli_result}). Contrary to \citet{ebrahimi-kann-2021-adapt}, we find that the best language adaptation strategy is Embedding-and-Adapters and the worst is Embedding-only. The difference can be explained by us constraining the continual training to the embedding layers only instead of the entire model, which makes the model harder to adapt to new languages. Therefore, the NLI performance increases as we add trainable parameters from invertible and language adapters. 
 
We also observe that it is easier to adapt to German, where adapted models~\modelref{bs_emb}--\modelref{bs100500_emb_adapt} almost close the gap with the upper bound baselines (\modelref{mbert}--\modelref{xglm}), but this is not the case for Korean, where this gap is still important.  This is in line with  \citeposs{muller-etal-2021-unseen} findings suggesting that knowledge transfer is more efficient between related languages. %about cross-lingual transfer learning.

\paragraph{Original Pretraining Steps.} The models \modelref{bs_emb_adapt}, \modelref{bs100500_emb_adapt} and \modelref{bs12000_emb_adapt} compare the effect of the pretraining steps on the new language adaptability.  We observe that, after language adaptation, the early pretraining checkpoint (row \modelref{bs12000_emb_adapt} in Table~\ref{tab:nli_result}) only performs as good as a non-adapted model \modelref{bs} when evaluated with prompts. However, after further finetuning on the downstream task, its performance catches up with later checkpoints. This result suggests that the model is capable to recover a lot of task knowledge during finetuning. Whether the knowledge encoded in the pretrained model was useful needs further investigation.

\paragraph{Adapters' Capacity.} The models with different adapters' capacity (models \modelref{bs_emb_adapt}, \modelref{bs_emb_adapt48} and \modelref{bs_emb_adapt384}) achieve comparable downstream performance when adapting to German and Korean languages. The small impact of parameter budget on the final performance could be due to the limited amount of training data. This implies that in the real low-resource settings there is no need for a very large parameter budget.

% when adapting to German language. We do however see that adapter layers (even small) boost performance for German. The Korean results are less conclusive though.  

\paragraph{Positional Embeddings.} Our results (models~\modelref{bs_emb_adapt} vs \modelref{bs_de_embpe_adapt}, models~\modelref{bs100500_emb_adapt} vs \modelref{bs_ko_embpe_adapt}) highlight the importance of adapting the positional embedding layer of the LM. For German, the adapted positional embeddings gives 6.5\% accuracy boost for cross-lingual NLI; for Korean, we obtain 4.9\% accuracy improvement for cross-lingual NLI and 12.7\% improvement for the supervised setting\footnote{The training of model \modelref{bs_de_embpe_adapt} diverged on Korean dataset, therefore we used a different original pretraining checkpoint to train \modelref{bs_ko_embpe_adapt}. We would investigate this issue further.}. This is in line with the findings by \citet{ravishankar-sogaard-2021-impact}.

%% file: sections/50_limitations.tex
\section{Limitations and Future Work}
\paragraph{Languages} In our language adaptation setup, we use pseudo-low resource languages by limiting the number of samples for German and Korean, which are classified as high-resource languages by \citet{joshi-etal-2020-state}. In our future work, we plan to explore language adaptation to truly low-resource languages such as Sinhala and Burmese.

\paragraph{Evaluation Tasks} We only evaluate our models on the natural language inference task, which is a high-level NLP classification task. We plan to extend our evaluation task suite to include generative tasks such as multilingual abstractive summarization \cite{hasan-etal-2021-xl} and sequence labeling tasks such as named entity recognition \cite{rahimi-etal-2019-massively} We acknowledge that the performances might not generalize to truly low-resource languages because they have exceptionally limited labeled resources.

\paragraph{Training Examples} We adapt the model to a new language by training on 100,000 samples of the new language. In reality, many languages may have far more monolingual data available. Our future experiments would include language adaptation with varying number of training samples and evaluate the effect of larger training samples on downstream performance and the use of adapters.

\paragraph{Origianl Pretraining Steps} As we observe that the model still performs well on downstream tasks after finetuning despite only 12,000 of original pretraining steps, we plan to evaluate how important it is to start from the pre-trained model by comparing the performance with a randomly initialized pretrained model. 

\paragraph{Tokenizers} We retrain the BigScience tokenizer with its original vocabulary size of 130,000 on the entire subcorpora of the new language. In our future work, we plan to explore the effect of the vocabulary size of the tokenizers and the number of samples used to train the tokenizers.

\section{Conclusion}
In this work we've compared different adaptation strategies to add a new language into pretrained BigScience model and identified important components for adaptation.

% \yzx{zero-shot task adapters}

% In addition, it would be interesting to include more intermediate pretraining checkpoints and characterize the relationship between the number of pretraining steps, the language-independent representations, and the ease of language adaptations.

%% file: sections/acknowledgments.tex
\section*{Acknowledgements}
The authors would like to thank anonymous reviewers for their helpful feedback.

%% file: sections/100_appendix.tex
\section{BigScience Pretraining} \label{appendix:bigs-pretraining}
The BigScience multilingual autoregressive language model (LM) uses the GPT architecture and the causal language modeling objective. Its tokenizer is the byte-level BPE with a vocab size of 130,001. For pretraining, it uses the Adam optimizer with a batch size 512, a learning rate of 2e-4 with cosine decay over 16,927,083 samples and warmup over 216,320 samples, weight decay of 0.1, and gradient clipping of 1.0. 

Table~\ref{tab:bigscience-pretraining-oscar} reports the statistics of the deduplicate subcorpora from OSCAR \citep{ortiz:oscar_2019} used for pretraining the BigScience model.

\begin{table}[ht]
\small
    \centering
    \begin{tabular}{lrr}
        \toprule
        Language & Number of Words & Sampling Probability \\
        \midrule
        Indonesian & 2,394,957,629 & 0.0554 \\ 
        Basque & 45,359,710 & 0.0184 \\
        Vietnamese & 5,577,159,843 & 0.0684 \\
        Chinese & 6,350,215,113 & 0.1339 \\
        Urdu & 218,030,228 & 0.0267 \\
        Spanish & 218,030,228 & 0.1118 \\
        Catalan & 729,333,440 & 0.0395 \\
        Portuguese & 10,751,156,918 & 0.0867 \\
        French & 23,206,776,649 & 0.1110 \\
        English & 215,841,256,971 & 0.2107 \\
        Hindi & 745,774,934 & 0.0398 \\
        Arabic & 3,171,221,354 & 0.0638 \\
        Bengali & 363,766,143 & 0.0339 \\
        \bottomrule
    \end{tabular}
    \caption{Statistics for OSCAR deduplicated subcorpora of 13 languages.}
    \label{tab:bigscience-pretraining-oscar}
\end{table}

\section{Language Adaptation Training Details}
\label{appendix:language-adaptation-training}
For all the language adaptation strategies, we train by using the AdamW optimizer with default parameters, batch size of 8, learning rate of $1e-3$ with linear decay, and maximum sequence length of 1,024. It takes around 30 hours to complete the training on a single V100 GPU machine.

\section{Relationship between NLI Task Performance and Adapter Capacity}
\begin{figure}[!ht]
\includegraphics[width=8cm]{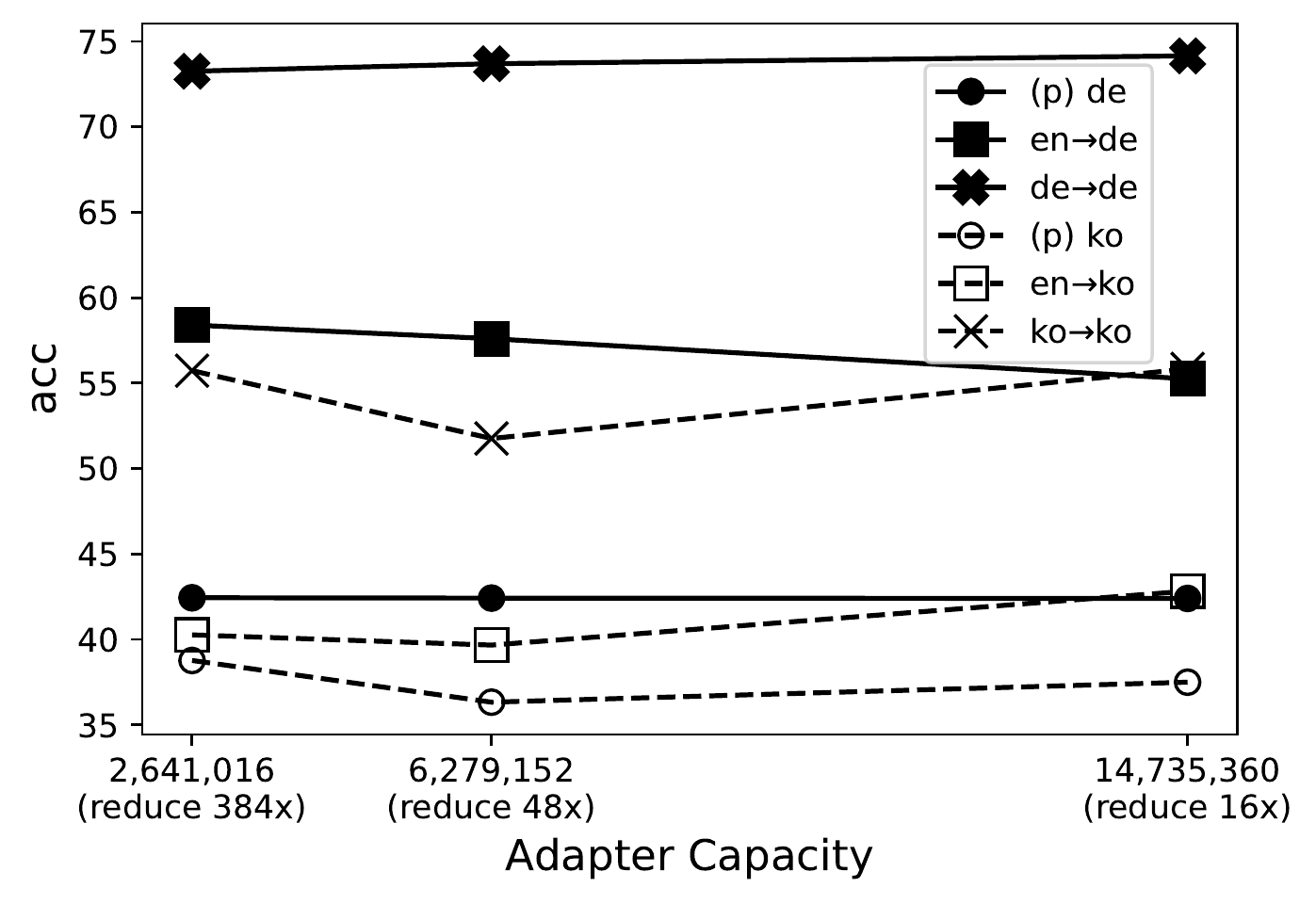}
\caption{Graph of test accuracy for NLI against total adapter capacity (after applying reduction factor) using the Embedding-and-Adapters adaptation.}
\label{fig:nli-adpt-capacity}
\end{figure}

We convert the reduction factor of language adapters to total adapter capacity and plot the NLI performance against it in Figure~\ref{fig:nli-adpt-capacity}. The figure shows comparable performance across all NLI evaluation settings for both German and Korean languages even with small adapter capacity.

